\documentclass[a4paper]{article}

\usepackage{amsthm, amsmath, amsfonts, amssymb}
\numberwithin{equation}{section}
\usepackage{mathtools}
\usepackage{mathrsfs}
\usepackage{enumitem}
\usepackage{hyperref}
\hypersetup{
    colorlinks,
    linkcolor={blue!30!black},
    citecolor={blue!50!black},
    urlcolor={blue!80!black}
}

\usepackage{color}
\usepackage[dvipsnames]{xcolor}
\usepackage{algorithm}
\usepackage{algorithmic}
\usepackage[normalem]{ulem}
\usepackage{ifthen}
\usepackage{pgf}
\usepackage{dsfont}
\usepackage{tikz}
\usetikzlibrary{arrows.meta}
\usepackage[T1]{fontenc}
\usepackage[utf8]{inputenc} %

\theoremstyle{remark}

\newif\iflong %
\longfalse%

\newcommand{\mb}{\mathbf}

\usepackage{xspace}

\DeclareMathOperator*{\argmin}{arg\,min}
\DeclareMathOperator*{\argmax}{arg\,max}

\newcommand{\bm}[1]{\mathbf{#1}}

\newcommand{\R}{\mathbb{R}}
\newcommand{\rset}{\R}

\newcommand{\ind}{\mathds{1}} %
\newcommand{\un}{\ind}
\newcommand{\point}{\,\cdot\,}

\usepackage{xifthen}
\newcommand{\PP}[1][]{\ifthenelse{\equal{#1}{}}{\ensuremath{\mathbb{P}}}{\ensuremath{\mathbb{P}\left( #1 \right)}}}
\newcommand{\EE}[1][]{\ifthenelse{\equal{#1}{}}{\ensuremath{\mathbb E}}{\ensuremath{{\mathbb E}\left[ #1 \right]}}}

\newcommand{\norm}[1]{\left\lVert#1\right\rVert}

\newcommand{\sphere}{\mathbb{S}}

\renewcommand{\argmin}{\textup{argmin}}

\newcommand{\pinbl}[1][]{%
  \ifthenelse{\equal{#1}{}}{\ell_\tau}{\ell_{#1}}%
}

\newcommand{\risk}[1][]{%
  \ifthenelse{\equal{#1}{}}{\mathcal{R}}{\mathcal{R}_{ #1 }}%
}

\newcommand{\emprisk}[1][]{%
  \ifthenelse{\equal{#1}{}}{\widehat{\mathcal{R}}}{\widehat{\mathcal{R}}_{ #1 }}%
}

\usepackage[comma]{natbib}
\usepackage{url}
\usepackage{fullpage}
\usepackage{authblk}

\usepackage[most]{tcolorbox}

\definecolor{tolblue}{HTML}{6699CC}
\definecolor{lightbrown}{HTML}{EB811B}
\tcbset {
  base/.style={
    arc=0mm, 
    bottomtitle=0.5mm,
    boxrule=0mm,
    coltitle=black, 
    fonttitle=\bfseries, 
    left=2.5mm,
    leftrule=1mm,
    right=3.5mm,
    title={#1},
    toptitle=0.75mm, 
  }
}

\newtcolorbox{textbox}[1]{
  colbacktitle=tolblue!50!white,
  colframe=tolblue,
  base={#1}
}

\newtcolorbox{issues}[1]{
  colbacktitle=lightbrown!50!white,
  colframe=lightbrown,
  base={#1}
}

\title{\bf Extrapolation in Statistical Learning\\ with Extreme Value Theory}

\author[1]{Sebastian Engelke}
\author[2]{Nicola Gnecco}
\author[3]{Anne Sabourin}
\affil[1]{Research Institute for Statistics and Information Science, University of Geneva, Switzerland}
\affil[2]{Department of Mathematics, Imperial College London, United Kingdom}
\affil[3]{Université Paris Cité, Université Paris-Saclay, ENS Paris-Saclay, CNRS, SSA, INSERM, Centre Borelli, France}

\begin{document}

\maketitle

\begin{abstract}
Extreme value theory provides rigorous theory and statistical tools for extrapolation in machine learning, particularly in settings where traditional methods struggle due to data scarcity in the tails. A broad range of tasks benefit from these advances, including regression and classification beyond the training data, extreme quantile regression, supervised and unsupervised dimension reduction, generative artificial intelligence and anomaly detection. 
This review synthesizes recent developments in these fields 
at the intersection of statistical learning and extreme
value theory, with a focus on principled methods based on asymptotically 
motivated representations of the tail of univariate and 
multivariate distributions.
We consider different theoretical frameworks
for both asymptotically dependent and independent data
and discuss how they translate into efficient 
statistical methods for extrapolation to extreme regions.
By addressing both theoretical and practical aspects, we offer a comprehensive overview of the state-of-the-art in this quickly
evolving field, and identify promising directions for future research.
\end{abstract}

\section{Introduction}
\label{sec:intro}

Statistical learning encompasses methods and theory for a wide range of 
tasks such as regression and classification, dimension reduction and generative models \cite[e.g.,][]{Hastie2009,Anthony1999,Vapnik2000}  %
On the methodological side, this field is closely connected to machine learning and artificial intelligence (AI), but with a stronger focus on uncertainty quantification, interpretability, and mathematical understanding of the models. 
Applications of statistical learning typically 
feature high-dimensional predictors or response variables, complex dependence structures and large data sets.
Recent examples include AI weather models \citep[e.g.,][]{Lam2023}, medicine \citep{Rajkomar2019}, industry \citep{Jan2023}. %

Flexible machine learning methods 
trained by empirical risk minimization excel at interpolation, that is, making predictions and drawing conclusions 
in regions with a sufficiently high density of data from the training distribution $\mathcal P$.
This practical success is supported by universal approximation results \citep[e.g.,][]{Cybenko1989} and statistical guarantees \citep[e.g.,][]{Schmidt-Hieber2020,Bousquet2004,Lugosi2002a}, which typically assume distributions of predictors and/or response with compact supports or sub-Gaussian tails, 
and evaluation on a test distribution $\mathcal Q = \mathcal P$ that 
coincides with the training distribution.
This theory, or variations thereof, apply to a wide range of 
tasks such as regression, classification or generative learning
of high-dimensional distributions.

On the other hand, situations that 
require predictions at points beyond the training
 data range violate the underlying assumptions of the theory since, for instance, the predictor space is unbounded, responses are heavy-tailed, 
 or there is a test distribution shift $\mathcal Q\neq \mathcal P$.
In practice, the performance of machine learning methods then often quickly degrades. 
The reason for the extrapolation issues is that flexible models such 
as neural networks make almost no assumptions on the data-generating process. 
While this flexibility is a strength for interpolation, 
it results in arbitrary behavior wherever information is missing. Examples for these extrapolation challenges appear in essentially any 
application, since generalization to new regimes is often 
a key interest.
A related field in machine learning is domain adaptation \citep{Ben-David2010}, %
which aims to develop methods that 
perform well on a test distribution that differs from the training data.
The approaches typically require data from the test distribution \citep{Ben-David2006}, assume that 
the test support is contained in the training support \citep{Sugiyama2007}, or assume causal 
structures \citep{Christiansen2022}. 
The extrapolation scenarios we consider here differ from these
more classical situations as we only require samples from the training distribution
and consider test points $\bm x \in \mathbb R^p$ far from the training data.

\begin{textbox}{Extrapolation challenges in applications}
The challenge of extrapolation arises in essentially 
all applications of machine learning and AI. 
In the recent field of AI-driven weather forecasting, it has been
observed that data-driven models generalize less well than physical models to record-breaking weather events that are more extreme than any 
training sample \citep{Pasche2025, Sun2025, Zhang2025}.  
Similar issues are present in the prediction of protein fitness \citep{Freschlin2024}, 
large language models \citep{Srivastava2023}, or the long-standing problem of peak river flows \citep{Martel2025}.
\end{textbox}

\begin{figure}[t]
\centering
\begin{tikzpicture}

\node[anchor=south west, inner sep=0] (img1) at (0,0)
  {\includegraphics[scale=.3,page=1]{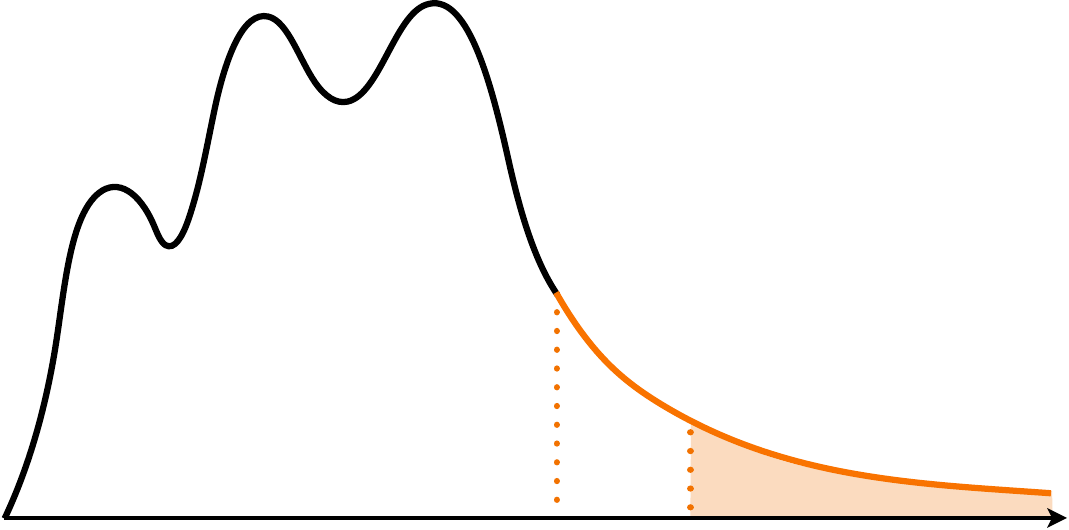}};

\node[anchor=south west, inner sep=0] (img2) at ([xshift=2.95cm]img1.south east)
  {\includegraphics[scale=.3,page=2]{figures/keynote-cropped.pdf}};

\begin{scope}[x={(img1.south east)}, y={(img1.north west)}, overlay]
  \node at (0.525,-0.05) {$u$};
  \node at (0.65,-0.05) {$x$};

  \node[align=left] at (0.95,0.65)
    {$\mathbb P(X \ge x)\propto
    \left(1+\gamma\dfrac{x-u}{\sigma}\right)_{+}^{-1/\gamma}$};

  \draw[-{Stealth}] (0.75,0.55) -- (0.80,0.20);
\end{scope}

\begin{scope}[
x={(img2.south east)},
y={(img2.north west)}, 
overlay]
  \node at (0.4,0.89) {$A$};
  \node at (0.46,0.49) {$tA$};
  \node at (0.75,0.25) {$\PP(\bm X \in A) \approx c(t)\PP(\bm X \in tA)$};
\end{scope}
\end{tikzpicture}
\caption{Left: Density of univariate $X$ with its right tail approximated by a generalized Pareto distribution. Right: Scatter of independent realizations of a random vector $\bm X \in \mathbb{R}^2$. The probability of the rare event $A \subset \mathbb{R}^2$ is approximated by the probability of the rescaled event $tA$, $t \in (0, 1)$, using a stability property of the multivariate tail of $\bm X$.}
\label{fig:diagram}
\end{figure}

The general difficulty of extrapolation from a non-parametric perspective is that there is no information in regions without data, and any statistical guarantee will require additional assumptions \citep{Stone1977}.
Extreme value theory provides suitable mild assumptions, the mathematical foundation, and the 
statistical tools to improve the extrapolation performance of 
machine learning methods \citep[e.g.,][]{DeHaan2007, Resnick2008}.
Similar to the central limit theorem, the tail of a
univariate response $X$ above some threshold $u$ can be
approximated under weak conditions 
by the generalized Pareto distribution, i.e.,
\begin{equation}\label{GPD_intro}
        \mathbb P(X \leq x \mid  X > u) \approx 1- \left(1+\gamma \frac{x - u}{\sigma}\right)_+^{-1/\gamma} , \quad  x \geq u,
\end{equation}
where $\sigma >0$ is a scale parameter, and the shape parameter $\gamma\in\mathbb R$ determines the 
heaviness of the tail of $Y$ \citep{Balkema1974}; see Figure~\ref{fig:diagram} (left)
for an illustration.
Estimation of these parameters on a training sample of size $n$ 
relies only on the $k\ll n$ exceedances over $u$, where $k$
is a tuning parameter that governs the bias-variance trade-off.
This extrapolation allows us to transfer information from moderately large training observations to regions with very few or no data points,
and it can improve machine learning
methods for tasks such as extreme quantile regression with predictors
or anomaly detection in possibly high-dimensional data sets. 

For a multivariate random vector $\bm X\in\mathbb R^p$, extrapolation
involves the dependence structure between the largest observations of the marginal components.  
Similarly to the univariate case, there exists asymptotic theory that shows that the structure in the multivariate tail simplifies compared to the bulk of the distribution.
The different approaches, including multivariate 
regular variation \citep{Resnick2007}, the conditional extremes model \citep{Heffernan2004} 
and the geometric approach \citep{Nolde2022}, all leverage a stability 
that allows to compute probabilities $\mathbb P(\bm X \in A)$
of rare events $A\subset \mathbb R^p$ from a shifted 
version $\mathbb P(\bm X \in t A)$, $t \in (0,1)$, that is less extreme; see Figure~\ref{fig:diagram} (right). 
The specific scaling factor $c(t) >0$ between the two probabilities depends
on the underlying framework.
In general, in the multivariate case, there is no longer a simple
parametric family as in Equation~\ref{GPD_intro} that describes
all possible tail behaviors, but the dependence structure
is typically non-parametric.
The field has developed effective tools
that are applied to improve extrapolation
in regression and classification tasks
or to help generative AI methods to
sample more accurately beyond the training 
data range.

For a training sample of size $n$, estimators from extreme value theory only use a smaller effective sample size $k\ll n$ of 
observations that are relevant for extrapolation. This 
exacerbates the curse of dimensionality and makes dimension reduction and high-dimensional methods even more relevant.
A lot of the recent work in multivariate
extreme value theory concentrates
on enforcing sparsity or reducing the dimensionality
in high-dimensional problems \citep{Engelke2021}. 
From the perspective of statistical learning theory, 
    recent work has demonstrated that the apparent obstacles to developing non-asymptotic learning guarantees for heavy-tailed data are not insurmountable. By leveraging the specific structure in the distributional tail provided by multivariate extreme value theory,  extremal distributions can be studied on bounded sets and 
    it is possible to derive non-asymptotic guarantees that scale with the effective sample size $k$ rather than the total sample size $n$ \citep{Lhaut2022,Clemencon2023}; see also the dedicated review \cite{Clemencon2025} for details and further references. %

In this article, we review recent works that build on principles 
from extreme value theory to address extrapolation
issues in complex statistical learning problems.
We include both methodological contributions that focus on a broad applicability to diverse data sets and approaches that derive theoretical guarantees under stronger assumptions. %
The paper is structured according to the different tasks that 
appear in statistical learning.
In Section~\ref{sec:supervised}, we consider two types of extrapolation challenges in supervised learning. First, in Section~\ref{sec:extreme_pred}, we discuss the case
where predictors $\bm X$ are more extreme than in the
training sample and both regression and classification methods must rely on extrapolation principles to guarantee reliable prediction.
Second, Section~\ref{extreme_QR} is concerned with quantile regression for quantile levels that exceed the data range of the response variable $Y$. Stable extrapolation in this case has been studied in the field of extreme quantile regression, combining univariate extreme value approximations with flexible machine learning methods.  
Section~\ref{sec:unsupervised} reviews literature on unsupervised learning. 
We focus on extreme value approaches that improve extrapolation of 
generative AI methods in Section~\ref{sec_genAI}, and that are designed 
to detect anomalies in multivariate data in Section~\ref{sec:anomaly}.
We also revisit the field of sparsity and dimension reduction in multivariate
extremes in Section~\ref{sec:dimRed}, which has seen significant progress since the review \cite{Engelke2021}.

\section{Background}\label{sec:background}

\subsection{Univariate extreme value theory}

Univariate extreme value theory provides the mathematical foundation to extrapolate the distribution of a random variable $X$ beyond the data range in terms of its tail probabilities or high quantiles.
In this section, we recall the main asymptotic results that 
motivate the limiting models used by many methods throughout this review.

There are two main approaches to describing the extremes of 
a random variable $X$ with distribution function  $F$ and upper endpoint $u^* \in (-\infty, \infty]$.
First, the peaks-over-threshold method describes the distribution 
of exceedances over a high threshold $u$ as it tends to the 
upper endpoint $u^*$. 
We then say that $F$ is in the domain of attraction of a generalized Pareto distribution 
with shape parameter $\gamma \in \mathbb{R}$ 
if there exists a scale function $\sigma :\mathbb{R}\to (0,\infty)$ such that 
\begin{equation}
    \label{eq:domain:attraction:1}
\lim_{u \to u^*} \mathbb P(X - u \leq x \sigma(u) \mid  X > u)  = 1- (1+\gamma x)_+^{-1/\gamma} := H_\gamma(x), \quad  x \geq 0,
\end{equation}
where $H_\gamma$ is the distribution function of the generalized Pareto distribution (GPD) and $a_+ = \max(0,a)$ for $a\in\mathbb R$ \citep{Balkema1974}. When $\gamma = 0$ this should be understood as the limit as $\gamma \downarrow 0$, i.e., $H_0(x) = 1-\exp(-x)$.
The shape parameter plays an important role as it determines the heaviness of the tail of $X$, ranging from heavy tails for $\gamma > 0$, through light tails for $\gamma = 0$, and distributions with finite upper endpoint for $\gamma < 0$.

Importantly, a large majority of well-known distributions
satisfy the domain-of-attraction condition of Equation~\ref{eq:domain:attraction:1} for some $\gamma\in\mathbb R$. Inverting this approximation at a fixed threshold $u$ then yields a strategy to approximate high quantiles of $X$ above level $\tau_0 = F(u)$ by
\begin{align} \label{eq:quantile:tail:approximation}
Q(\tau) = F^{-1}(\tau) \approx u  + \frac{\sigma(u)}{\gamma} \left\{ \left(\frac{1-\tau_0}{1-\tau}\right)^{\gamma} - 1 \right\}, \quad \tau > \tau_0.
\end{align} 
In practice, to compute these extrapolations based on an independent sample $X_1,\dots, X_n$ of $X$,
 we first replace the threshold $u= X_{n-k:n}$ by the $(n-k)$th order statistic of the sample. This corresponds to the empirical quantile at level $\tau_0 = 1-k/n$, where $k$ is the effective sample size of exceedances and its choice represents a bias-variance tradeoff. We further obtain estimators $\hat \sigma = \hat \sigma(u)$ and $\hat \gamma$ of the GPD parameters, for instance by maximum likelihood estimation.

 The second approach in univariate extreme value theory studies the limit of block maxima $M_n = \max(X_1,\dots , X_n)$.
 For a random variable satisfying the domain-of-attraction condition of Equation~\ref{eq:domain:attraction:1}, there exist sequences $a_n >0$ and $b_n \in\mathbb R$ such that 
 \begin{align}\label{gev}
     \lim_{n\to\infty} \mathbb P\left( \frac{M_n - b_n}{a_n} \leq x \right) = \exp\left\{ - (1+\gamma x)_+^{-1/\gamma} \right\} := G_\gamma(x), \quad  x \geq 0,
 \end{align}
where the limit is called the generalized extreme value (GEV) distribution. For a fixed block size $n$, we can again
invert the approximation in Equation~\ref{gev} to obtain  
of the $\tau$-quantile of $X$ by $Q(\tau) \approx G_\gamma^{-1}\{1 - (1-\tau)/n\}$.

\subsection{Multivariate extreme value theory}\label{sec:mevt}

For a random vector $\bm X$ in $d\geq 2$ dimensions, multivariate 
extreme value studies the extremal dependence between large
observations in the different components. Naturally, this task
is much more involved than the univariate problem and 
there does not exist a simple parametric family that describes
all possible tail behaviors. Instead, there are several different
modeling approaches to describe extremal dependence. 
In particular, the extremal dependence between two components $X_i$
and $X_j$ with distribution functions $F_i$ and $F_j$ can be categorized broadly into two different regimes
according to their extremal correlation (or upper tail dependence) coefficient
\begin{align}\label{eq:chi}
 \chi_{ij} = \lim_{q\to 1}\mathbb P(F_i(X_i)>q,F_j(X_j)>q) / (1-q) \in[0,1],
\end{align}
whenever the limit exists. If $\chi_{ij} > 0$, then we speak of
asymptotic dependence, and if $\chi_{ij}=0$ we say that the two
components are asymptotically independent.

The most classical framework for asymptotic dependence is multivariate regular variation \citep[][Section 5]{Resnick2008}.
There are various ways of introducing this notion, and here 
we choose the formulation for threshold exceedances, which is closest to the univariate case of Equation~\ref{eq:domain:attraction:1}.
To simplify notation and to abstract away from marginal distribution,
we assume %
that the marginal
distributions of $\bm X$ are standard Pareto, which can be ensured in practice by preliminary marginal transformations based on probability integral transforms.
The random vector $\bm X$ is multivariate regularly varying
if its rescaled threshold exceedances converge in distribution to a multivariate Pareto distribution $\bm X_\infty$, that is, 
\begin{align}\label{eq:MPD}
	\mathbb P(\bm{X}_\infty \le \bm x)=\lim_{u\to\infty}\mathbb P(\bm X/u \le \bm x \mid \max_{i=1,\dots, d} X_i >u), \quad \bm x\in\mathcal{L}, 
\end{align}
where $\mathcal{L}=\{\bm x\in\mathbb R^d: \max_{i=1,\dots, d} x_i >1\}$ \citep{Rootzen2006}.
Alternatively, multivariate regular is often framed in polar coordinates. 
For some norm $\|\cdot\|$, we define the radial and 
angular components as $R = \|\bm X\|$ and 
$\mb W =   \bm X / R$, respectively.
The convergence in Equation~\ref{eq:MPD} is equivalent to 
 \begin{equation}\label{RTheta_cv}
     \PP( R / t > s, \mb W \in B\mid R >t ) \to s^{-1} \Phi(B), 
 \end{equation}
 where $\Phi$ is the spectral measure on the unit sphere $\sphere_{d-1}$,
 and $B\subseteq \sphere_{d-1}$ is a Borel subset.
In other words, conditioning on a large radius $R$, the distributions
of the radius and the angle are approximately independent. 
The latter can be any distribution $\Phi$ on the sphere as long as 
it satisfies certain moment constraints stemming from the choice of standard Pareto margins \citep[e.g.,][]{Einmahl2009}. A third interpretation of multivariate regular
variation is in terms of componentwise
maxima of i.i.d.~copies of $\bm X$, leading to the multivariate
version of the GEV distribution \citep{deHaan1977}.

For the remainder of this section, we assume that $\bm X$ is 
normalized to standard Laplace margins, since approaches for asymptotic independence are more naturally expressed in 
light-tailed margins.
A generalization of multivariate regular variation that also 
covers asymptotic independence is the 
conditional extremes model introduced in \cite{Heffernan2004}.
For this approach, we single out $X_j$ for some $j\in\{1,\ldots, d\}$ and condition on large values of this component. With normalization given by the two 
vectors $\bm a^{(j)} \in (0,1]^{d-1}$ and $\beta^{(j)}\in (0,1)^{d-1}$, 
the conditional extremes model assumes the existence of a non-degenerate 
limit $\bm Z$ in dimension $d-1$ such that
\begin{align}\label{eq:HT}
	\mathbb P(\bm Z \le \bm z)=\lim_{u\to\infty}\mathbb P(\bm X_{-j} \le \bm  a^{(j)} X_j + X_j^{\beta^{(j)}} z \mid  X_j>u), \quad \bm z\in\mathbb R^{d-1}; 
\end{align}
for more general normalizations we refer to \cite{Heffernan2004}. 
In particular, for the case of asymptotic dependence under multivariate regular variation, this assumption is satisfied with $\bm a^{(j)} = \mathbf{1}$ and $\beta^{(j)} = \mathbf 0$. More generally, for indices where $a^{(j)}_{\ell} < 1$ the components $Y_\ell$ and $Y_j$ are asymptotically independent.

Another approach to describe extremal dependence of a random vector with density $f_{\bm X}$ is through a geometric perspective \citep{Balkema2010, Nolde2022}. To this end, define the gauge function as
\begin{align}
    \label{gauge} g(\bm x) := \lim_{t\to\infty} -\log f_{\bm X}(t\bm x) /t, \quad \bm x \in \mathbb{R}^d,
\end{align}
whenever the limit exists.{for me}
It can be shown that the latter implies that the rescaled sample cloud 
$\{\bm X_1/\log n, \dots, \bm X_n /\log n\}$ converges onto the limit 
set $G = \{\bm x\in\mathbb R^d: g(\bm x) \leq 1\}$.
These objects can be used to describe the extremal dependence in
many known distributions, in particular those that exhibit asymptotic independence.

\section{Extrapolation in Supervised Learning}\label{sec:supervised}

\subsection{Setting}

Supervised learning is concerned with the prediction of a response
variable $Y$ given a predictor vector $\bm X$ with values in some set $\mathcal X \subseteq \mathbb R^p$.
The response variable typically takes values in $\mathcal Y = \mathbb R$ (or possibly in $\mathbb R^d$) in which case we speak of regression, or in a finite set $\mathcal Y$ and we speak of classification.
The goal is to construct a prediction model $f: \mathcal X \to \mathcal Y$ to estimate some statistic of the conditional distribution of $Y\mid \bm X = \bm x$ for a new test predictor $\bm x \in \mathbb R^p$ of interest. 
The performance of such a model is measured by specifying a loss function $\ell: \mathcal Y \times \mathcal Y \to \mathbb R$ and considering
the expected risk
\begin{align}\label{def_risk}
    R(f) = \mathbb E_{\mathcal Q}\left[ \ell\{ f(\bm X), Y\}\right],
\end{align}
where the expectation is taken over a test distribution $\mathcal Q$ of $(\bm X, Y)$.
The minimizer $f^*$ of Equation~\ref{def_risk} is called the Bayes predictor
that attains the Bayes risk $R(f^*)$.
In classification, the most common loss function is 0--1 loss $\ell(y,z) = \un_{\{y \neq z\}}$ and the corresponding Bayes predictor
$f^*(\bm x)=\argmax_{z \in \mathcal Y} \mathbb P_{\mathcal Q}(Y = z \mid \bm X = \bm x)$.
In regression, often the squared error $\ell(y,z) = (y-z)^2$ is used, which implies the conditional mean
$f^*(\bm x) = \mathbb E_{\mathcal Q}(Y \mid \bm X = \bm x)$ as Bayes predictor. 
While the squared error loss focuses on the center of the
response distribution, quantile regression uses the quantile check function $\ell(y,z) = (y-z)(\tau-\un_{\{ y<z\}})$ \citep{Koenker1978} at some
probability level $\tau \in (0,1)$. The Bayes predictor is then
the $\tau$-quantile $f^*(\bm x) = q_{\bm x}(\tau)$ of the conditional distribution of $Y \mid \bm X = \bm x$ under $\mathcal Q$. In particular, for $\tau = 1/2$ this reduces
to the absolute error with $\text{median}(Y \mid \bm X = \bm x)$ as Bayes predictor.
For more details on supervised learning see, \emph{e.g.}, the monographs~\cite{Vapnik2000,Devroye2013,Bousquet2004}. %

Given a training sample $(\bm X_i, Y_i)$, $i=1,\dots, n$, generated from the training distribution $\mathcal P$ of $(\bm X, Y)$, a prediction model approximates the Bayes predictor by minimizing the empirical risk (or training error) 
\begin{align}\label{def_emp_risk}
    \widehat R(f) = \frac1n  \sum_{i=1}^n \ell\{ f(\bm X_i), Y_i\}.
 \end{align}
Establishing theoretical and empirical guarantees regarding this approximation may be seen as the core idea of statistical learning \citep[e.g.,][]{Schmidt-Hieber2020,Bousquet2004,Vapnik2000}. Such guarantees typically require that the test distribution $\mathcal{Q}$ be identical to the train distribution $\mathcal{P}$, with bounded sample spaces or sub-Gaussian tails. 

In machine learning, a well-known challenge is the 
out-of-distribution generalization problem where the test points 
are sampled from a different distribution $\mathcal Q$ than the training distribution $\mathcal P$.
Test predictors are then more likely to lie outside of the training data
range so that they require extrapolation.
While machine learning methods are very good at interpolation, they notoriously struggle if the test predictor is more
extreme (in some sense) relative to the set of training predictors.
This extrapolation problem is linked to domain adaptation and generalization, which aim to develop models that 
perform well on a test distribution that differs from the training data \citep{Ben-David2006, Sugiyama2007, Rothenhausler2021, Christiansen2022}.

The extrapolation scenarios we consider here differ from these
more classical situations as we only require samples from the training distribution
and consider test points $\bm x \in \mathbb R^p$ far from the training data, %
that are seen as realizations from the same distribution as the training data. Occurrences of other events at least as large in some sense (say $\{\|X\|>\|x\|\}$ for some norm $\|\point\|$) have low probability, but are bound to happen in the long run, a typical motivation in risk analysis. Thus, our framework involves identical train and test distribution $\mathcal{P} =\mathcal{Q}$, although the focus is on the tails of $\mathcal{P}$ (in a sense that depends on the specific context), while the bulk plays practically no role. %
In Section~\ref{sec:extreme_pred} we discuss methods that use extreme value theory to tackle this problem, both in regression and classification, where $\mb X$ represents the predictor (or covariate vector) and the focus is on unusually large values of the predictor.

\begin{figure}[t]
    \centering
    \includegraphics[scale=0.6]{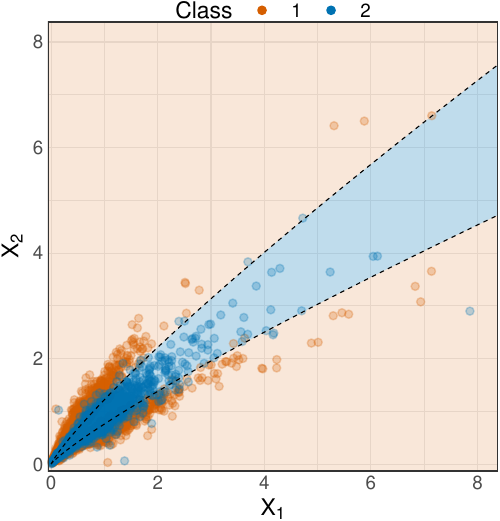}
    \includegraphics[scale=0.6]{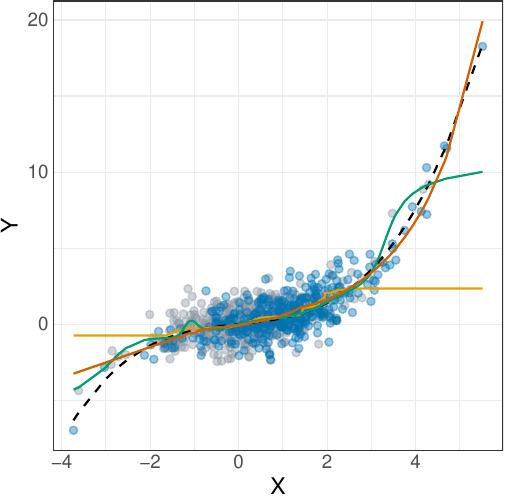}
    \includegraphics[scale=0.6]{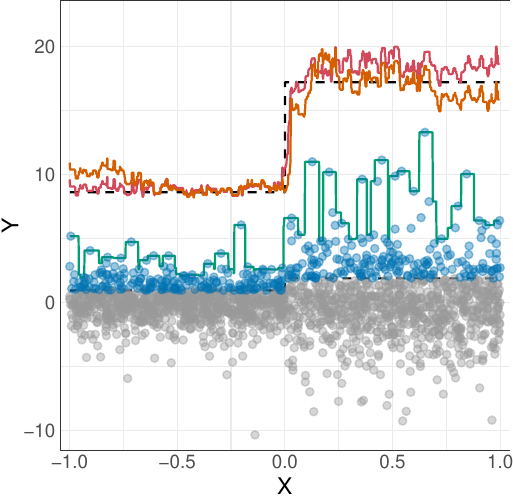}
    \caption{Left: Extrapolation in classification when $\bm X$ can take extreme values,  based on angular components of extremes. Marginal distributions have been transformed to Pareto$(\alpha= 3)$ for visual clarity.
    Center: Extrapolation in regression when $X \in \rset$ can take extreme values. 
    The dashed line represents the true regression function; the red line represents a fitted function based on an extrapolation-aware method; the yellow and green lines represent functions fitted by a random forest and a neural network, respectively.
    Grey and blue points represent training and test observations, respectively.
    Right: Extrapolation in quantile regression when $Y$ can take extreme values, conditional on $\bm X$.
    Grey (blue) points represent training observations below (above) an intermediate quantile.
    The dashed line represents the true $99.95\%$ conditional quantile.
    Orange and red lines represent fitted quantile curves using methods based on extreme value theory. The green line represents a fitted quantile curve using random forests, which lack extrapolation capabilities.
    }
    \label{fig:classification-regression}
\end{figure}

A second type of extrapolation occurs in supervised settings  because of data scarcity 
in the direction of the response variable~$Y$.
Predicting a conditional quantile $q_{\bm x}(\tau)$ accurately
using the empirical risk with the quantile loss requires
sufficient samples in a neighborhood of $\bm x$ that exceed this
quantile. For extremely high or low quantiles where $\tau \to 1$ or $\tau \to 0$, respectively, classical quantile regression methods fail.
Such extreme quantile regression
is a classical task in extreme value theory and has recently been 
combined with flexible machine learning methods to interpolate
in the predictor space. In Section~\ref{extreme_QR} we 
present different approaches %
to extreme quantile regression and discuss related topics such as the prediction of extreme events and dimension reduction guided by extreme responses.
Figure~\ref{fig:classification-regression} summarizes the main extrapolation problems considered in this section, namely classification and regression in extrapolation regimes of the covariate (left and center panel) and extreme quantile regression (right panel).

\subsection{Extreme Predictors}\label{sec:extreme_pred}

\subsubsection{Classification}\label{sec:classif}

We consider binary classification where the response $Y$ takes values
in the set $\mathcal Y = \{0, 1\}$ and the loss function is 0--1 loss $\ell(y,z) = \un_{\{y\neq z\}}$. The risk $R(f)$ defined in Equation~\ref{def_risk}
evaluates the performance of some classifier $f:\mathcal X \to  \mathcal Y$ under the test distribution $\mathcal Q$.
The Bayes classifier minimizing this risk is given by $f^*(\bm x) = \un_{\{\eta (\bm x) \geq 1/2 \}}$, where $\eta(\bm x) = \PP_{\mathcal Q}(Y = 1 \mid \bm X = \bm x)$. 
In this section, we assume that the training and test distributions
coincide $\mathcal Q = \mathcal P$, and we drop any subscripts
for simplicity. We further assume that the marginal distributions of $\bm X$ are normalized to standard Pareto distributions. To explicitly assess the extrapolation behavior of the classifier,
\cite{Jalalzai2018} propose a tail risk by $t>0$ by restricting
the area of interest to large values of the predictor in terms
of some norm $\|\cdot \|$,
\[
\quad R_t(f) = \PP\{Y \neq f(\bm X) \mid \|\bm X\| > t\}, 
\]
In order to define a limiting risk functional $R_\infty$ as $t\to \infty$, we 
need to restrict the possible test distribution $\mathcal P$.
We assume that both class distributions $\bm X \mid Y = 0$ and $\bm X \mid Y = 1$
are multivariate regularly varying as in Equation~\ref{eq:MPD}, and
that class probabilities $\PP(Y = \sigma \mid \|\bm X\| > t)$, $\sigma \in \{0,1\}$,
are asymptotically comparable in the sense that their ratio converges to
a non-zero constant.
Under these assumptions, the rescaled random vector $(\mb X/t, Y)$, conditioned on $\|\bm X\|>t$, converges in distribution  as $t\to\infty$,   to a limit $(\mb X_\infty, Y_\infty)$ 
which describes the relation between extreme values
of the predictor and the response. 
The limiting risk functional is then defined as
\[ R_\infty(f) = \limsup_{t\to\infty} R_t(f).\]
Under additional technical assumptions, \cite{Jalalzai2018} show that the limiting Bayes classifier with respect to $R_\infty$ is given by $f_\infty^*(\bm x) = \un_{\{\eta_\infty (\bm x) \geq 1/2 \}}$, where $\eta_\infty(\bm x) = \PP(Y_\infty = 1 \mid \bm X_\infty = \bm x)$, and it only
depends on the angle, i.e., $f^*_\infty(\bm x) = f^*_\infty\circ w(\bm x)$
with $w(\bm x) = \bm x/\|\bm x\|$ and for all $\bm x\in \rset^p\setminus\{0\}$.
Importantly, 
\[ R_t(f_\infty^*) - R_t(f^*) \to 0, \quad t \to \infty,\]
that is, the limiting Bayes classifier attains a pre-asymptotic risk
that becomes arbitrarily close to the Bayes risk in extreme regions. 
This means that one may train a classifier over the simpler class of angular functions $f(\mb x)= h\circ w(\mb x)$ 
and use their invariance property for extrapolation.
In practice, given the training sample $(\bm X_i, Y_i)$, $i=1,\dots, n$, 
from the training distribution $\mathcal P$,
this can be achieved through empirical risk minimization using only the  $k\ll n$ largest observations, %
\begin{equation}\label{erm_angular}
    \widehat{R}_k(h) = \frac{1}{k} \sum_{i=1}^k \un\{Y_{(i)} \neq h\circ w (\bm X_{(i)})\},
\end{equation}
where \( (\bm X_{(1)}, Y_{(1)}), \ldots, (\bm X_{(n)}, Y_{(n)}) \) are sorted in
descending order in \( \|\bm X_i\| \). 
The minimization is performed over a family of 
angular classifiers $h\circ w $ with controlled complexity with finite Vapnik--Chervonenkis dimension, a key concept in statistical learning theory for quantifying model capacity \citep[e.g.,][]{Lugosi2002a}.
The effective sample size $1\leq k\leq n$ 
governs how far in the tail the classifier is trained and is a tuning parameter for the bias-variance trade-off, as illustrwated in the left panel of Figure~\ref{fig:classification-regression}. 
 Theoretical guarantees in terms of finite-sample bounds for the empirical risk minimizer 
are established in \cite{Jalalzai2018}, with extensions to predictors with different marginal distributions via rank transformations in \cite{Clemencon2023}; see also the review \cite{Clemencon2025} for more details.
\cite{Jalalzai2020} apply the above classification method in extreme regions to natural language processing, leveraging a new 
heavy-tailed word embedding.

Cross-validation (CV) for model selection in this context is addressed from a statistical learning theory perspective in \cite{Aghbalou2024}.
Error bounds are derived for CV estimates of the risk of a classifier in a low-probability region, paralleling so-called sanity-check bounds from the statistical learning literature \citep{Cornec2017}. These bounds provide guarantees for model selection or hyperparameter tuning. 
For instance, in constrained logistic regression on extreme covariates, the predictor takes the form $f_\beta(\bm x) = \textrm{sign}(\langle \beta, w (\bm x) \rangle)$, where $\beta \in \mathbb{R}^p$ and $\|\beta\|_1 \leq C$ for some constant $C$ that can be effectively chosen using CV.
The CV errors satisfy
bounds involving the effective sample sizes $pn_{V}$ and $pn_{T}$ of the training sets and validation sets, respectively, where $p$ is the probability of the rare region.

\subsubsection{Regression}

In regression, the usual modeling assumption is that the 
real-valued response is generated through
\begin{align}
    \label{reg_fun}
    Y = g(\bm X, \varepsilon),
\end{align}
where the regression function $g: \mathcal X \times \mathbb R \to \mathbb R$ depends on the predictor and an independent noise variable $\varepsilon$.
While this is the general framework, the most common assumption
is the additive noise model where $g(\bm X, \varepsilon) = f(\bm X) + \varepsilon$ for some function $f:\mathbb R^p \to \mathbb R$.

There are several ways to define which points require extrapolation,
and we first focus on the case of a one-dimensional predictor variable $X$ with values in $\mathcal X \subseteq \mathbb R$.
Following a population perspective, we can define an extrapolation point
as a data point $x \in \mathcal X$ that is not in the support of 
the training distribution, that is, $x \notin \text{supp}(\mathcal P_{X})$, where $\mathcal P_{X}$ denotes the marginal distribution
of the predictors under the training distribution $\mathcal P$.
\cite{Shen2024} develop a method to perform extrapolation 
in this setting under the assumption of a pre-additive noise model
where $g(X, \varepsilon) = f(X + \varepsilon)$ for some function $f:\mathbb R \to\mathbb R$, that is, the noise $\varepsilon$ is added on the level of the predictors. They prove out-of-distribution identifiably for one-dimensional predictors and propose the distributional regression method called engression
to leverage extrapolation under the pre-additive noise model.
In same framework of bounded predictor support, \cite{Pfister2024} assume that the $q$th derivatives of the conditional mean function $\mathbb E(Y\mid X = x)$ on out-of-support $x$ are bounded by those in-support.
They derive upper and lower extrapolation bounds for the worst-case
behavior of the regression function outside of training support.

In practice, the framework of bounded training domains might be too restrictive, since extrapolation
is often needed even if $x \in \text{supp}(\mathcal P)$. In fact, for a given sample size $n$, extrapolation is related to the range
of the training sample $(X_i, Y_i)$, $i=1,\dots, n$, since in regions
with few or no data points (even inside the support), non-parametric estimation becomes difficult. 
This motivates a data-centered perspective that is very common in extreme
value theory: the meaning of extrapolation (or extreme) 
is relative to the available data; see Section~\ref{extreme_QR} where
an extreme quantile level is defined to be above the highest data point.

Extreme value theory allows for assumptions that differ from 
the structural model assumption above and are more of
distributional type that can be verified for many model classes.
Similar to the power transformations of the response variable
in \cite{Box1964} that are designed to
simplify the relation with predictors (e.g., to make it linear), 
\cite{Buritica2024} propose to transform both predictor and response to Laplace margins $X^* = F_L^{-1}\circ F_X(X)$ and $Y^* = F_L^{-1}\circ F_Y(Y)$, where $F_L$ is the Laplace distribution function.
The dependence between $X^*$ and $Y^*$
in extreme regions where $X^*$ is large is well-understood in multivariate 
extreme value theory; see multivariate regular variation and the conditional extremes approach in Section~\ref{sec:mevt}. In particular, the conditional median can be shown to be close to linear in many cases:
\[ \text{median}(Y^* \mid X^* = x^*)= a x^* + (x^*)^\beta b + r(x^*),\]
where $a\in [-1,1]$, $b\in\mathbb R$, $\beta \in [0,1)$ and a remainder term that is asymptotically negligible.
Leveraging this, \cite{Buritica2024} propose an extrapolation principle based on the progression approximation of the conditional median and show that the relative error can be controlled
even outside of the data range up to a certain extrapolation limit 
that depends on the sample size $n$.
For the additive noise model $Y = f(X) + \varepsilon$, they
give conditions on the regression function $f$ and the tail heaviness of 
$X$ and $\varepsilon$ under which the progression extrapolation is
valid. Also for the pre-additive noise model from~\cite{Shen2024}, 
progression can be shown to be applicable.
The progression method is further extended to multivariate predictors $\bm X$ 
where the response follows an additive model.

\cite{Clemencon2025a} and \citet[][Section 5]{Clemencon2025}
propose a regression extrapolation approach 
for a bounded response $Y$ and squared-error
which is based on the assumption of multivariate regular variation as in Equation~\ref{eq:MPD}. Similar to the classification setup in \cite{Jalalzai2018},
they assume regular variation with respect to the predictors, namely that $(\bm X/t,
 Y)$, conditional on $\|\bm X\| > t$ converges in distribution to a limit $(\bm X_\infty, Y_\infty)$ as $t\to\infty$.  
They show that it is then sufficient to consider tail regression
 functions $h\circ w (\bm x)$ that depend only on the angle $w (\bm x)$ of the predictor
 when $\|\bm x\|$ is sufficiently large, and promote 
 learning via 
 empirical risk minimization as in Equation~\ref{erm_angular} but with squared-error loss.  
Their framework is applicable, for instance, when the response $Y$ arises from an appropriate rescaling of an initial target $Z$ for which the vector $(\mb X, Z)$ exhibits classical multivariate regular variation.

The theory in \cite{Clemencon2025a} is constrained to pure empirical risk minimization. Since most high-dimensional machine learning methods incorporate regularization, \cite{Clemencon2025} extend this line of work by analyzing a lasso-type algorithm within a linear class of angular predictors  $h_\beta\circ w (\mb x) = \langle \beta, w (\mb x)\rangle$ indexed by $\beta\in\rset^p$, but where the minimization problem includes an $\ell_1$-penalty $\|\beta\|_1$.
They derive non-asymptotic, high-probability bounds on the prediction error, analogous to classical results but with the sample size $n$ replaced by the effective sample size $k$. 
These bounds hold under the regression model 
\[
Y = f(\mb X)  + \langle \mb \beta, w (\mb X) \rangle + \sigma(\mb X)\varepsilon,
\]
that is flexible in the bulk of $\mb X$ yet linear in the tail,
where $f:\mathcal X \to \mathbb R$ is an arbitrary regression function 
that may dominate in the bulk but vanishes in the tail as $\|\bm x\| \to \infty$.
The quantity $\sigma:\mathcal X \to \mathbb R$ captures heteroscedastic noise variance
and is assumed to depend only on the angle $w (\mb x)$ for 
large $\|\bm x\|$. 
The tail regression function $h_\beta\circ w (\mb x)$ then only depends on the angle as required by the above theory.

\subsection{Extreme Response}\label{extreme_QR}

\subsubsection{Extreme quantile regression}

Given a pair $(\bm X,Y)$ of predictor and response, an important task is to estimate the conditional quantile $q_{\bm x}(\tau)$ of $Y \mid \bm X = \bm x$ at level $\tau \in (0,1)$. 
Quantile regression leverages the representation 
\begin{equation}\label{qloss}
q_{\bm{x}}(\tau) = \argmin_{q\in \mathbb R} \mathbb E[\rho_{\tau}(Y - q) \mid \bm{X} = \bm{x}],
\end{equation}
where $\rho_{\tau}(t) := t(\tau-1_{\{ t<0\}})$ is the quantile check function \citep{Koenker1978}.
For $n$ samples $(\bm X_i, Y_i)$, $i=1,\dots,n$, one estimates $q_{\bm{x}}(\tau)$ 
by minimizing the empirical quantile loss 
\begin{equation}\label{emp_qloss}
\hat{q}_{\bm{x}}(\tau) = \argmin_{q\in\mathcal{M}} \dfrac{1}{n}\sum_{i=1}^n \rho_{\tau}(Y_i - q (\bm{X}_i)),
\end{equation}
over a model class $\mathcal{M}$, e.g., 
linear models~\citep{Chernozhukov2005}, random forests~\citep{Athey2019}, or neural networks~\citep{Cannon2011}. 
Such methods 
perform well 
for moderate levels $\tau$, where enough observations exceed the target quantile.
However, for more extreme levels, 
the empirical minimization in Equation~\ref{qloss} can result in
a large bias \citep[see, e.g.,][Figure 3]{Pasche2024}.

To overcome the bias of empirical methods
for estimation of extreme quantiles, asymptotic approximations 
of the distributional tail motivated by extreme value theory are usually employed.
One line of work assumes $Y \mid \bm X = \bm x$ is heavy-tailed and, for $\tau \to 1$, extrapolates from an intermediate quantile level $\tau_0 < 1$ using the Weissman approximation
\begin{equation*}
{q}_{\bm x}(\tau)
\approx
{q}_{\bm x}(\tau_0)
\left(\frac{1-\tau_0}{1-\tau}\right)^{{\gamma}(\bm x)},
\end{equation*}
where $\gamma(\bm x)$ is the conditional shape parameter \citep{Weissman1978}.
\cite{Daouia2011} estimate ${q}_{\bm x}(\tau_0)$ by inverting the conditional survival function estimated with kernel methods, and estimate $\gamma(\bm x)$ with, e.g., the Hill estimator localized using kernels.
\cite{Gardes2019} propose an integrated version of this Weissman estimator.

A second line of work follows the peaks-over-threshold approach and assumes that $Y \mid \bm X = \bm x$ lies in the domain of attraction of a GPD.
First, an intermediate quantile function 
$\hat q_{\bm x}(\tau_0)$ is estimated with a classical quantile
regression method, and,
second, the exceedances
\begin{align}
    \label{exceedances}
    Z_i = (Y_i -  \hat q_{\bm X_i}(\tau_0))_+, \quad i=1,\dots, n,
\end{align}
are modeled 
by a conditional %
GPD as defined in Equation~\ref{eq:domain:attraction:1} with scale and shape parameters
$\hat \sigma(\bm X_i) > 0$ and $\hat \gamma(\bm X_i)\in \mathbb R$, respectively.
Conditionally on a predictor value of interest $\bm x \in\mathbb R^p$, this yields an extreme quantile estimator at extreme level $\tau > \tau_0$
\begin{equation*} 
          \hat q_{\bm x}(\tau) =  \hat q_{\bm x}(\tau_0)+\hat \sigma(\bm x) \frac{\left( \frac{1-\tau_0}{1-\tau}\right)^{ \hat \gamma(\bm x)}-1}{ \hat\gamma(\bm x)}.
\end{equation*}
The intermediate quantile $q_{\bm x}(\tau_0)$
together with the conditional GPD parameters $\vartheta(\bm x) = (\sigma(\bm x), \gamma(\bm x))$ characterize the tail of $Y \mid \bm X = \bm x$.
Earlier work estimate
$\vartheta(\bm x)$ through parametric forms such
as linear \citep{Chernozhukov2005, Wang2012} or additive models \citep{Chavez-Demoulin2005, Youngman2019}.
More recent approaches
model $\vartheta(\bm x)$ locally 
using machine learning methods. One possibility is to explicitly build a decision tree that places splits to maximize the GPD log-likelihood $\ell_\vartheta$ \citep{Farkas2021}. Another approach is to
maximize the weighted GPD log-likelihood 
\begin{equation}\label{eq:weighted-loglik}
  L_{n}(\vartheta; \bm x) =  \sum_{i =1}^{n} \omega_n(\bm x, \bm X_i) \ell_\vartheta(Z_i) 1\{Z_i > 0\}, \quad x \in \mathcal {X},
\end{equation}
where 
$\ell_\vartheta$ denotes the likelihood of the GPD model in Equation~\ref{eq:domain:attraction:1}, and
the weights $\omega_n(\bm x, \bm X_i)$ describe the 
similarity between the training predictor $\bm X_i$ and $\bm x$, and can be learned via
random forests \citep{Gnecco2024},
gradient boosting \citep{Velthoen2019, Koh2023},  neural networks \citep{Pasche2024, Richards2025}, or with high-dimensional statistical srategies such as Bayesian lasso \citep{deCarvalho2022}.
These methods combine the advantages of extreme value extrapolation of the response $Y$ with machine learning to handle large-dimensional predictors $\bm X$; see \cite{Tang2026} for a more detailed review.

One can interpret extreme quantile regression methods
as probabilistic forecasts 
that show better properties for extreme outcomes than classical approaches. 
To assess this, \cite{Allen2025} introduce 
a notion called tail calibration that particularly targets exceedances over a high threshold.
This framework is linked to the construction of prediction intervals with correct coverage.
For a confidence level $1-\alpha \in (0,1)$,
the (one-sided) prediction set $C_\alpha(\bm x) = (-\infty, q_{\bm x}(1 - \alpha)]$ satisfies the marginal coverage 
\begin{align}
    \label{marginal_cov}
    \mathbb P\{Y_{\text {test}} \in C_{\alpha}(\bm X_{\text {test}})\} \geq 1-\alpha, 
\end{align}
for a test point $(\bm X_{\text {test}}, Y_{\text {test}})$; in fact, it even satisfies the corresponding conditional coverage.
In practice, the quantile functions $q_{\bm x}(1-\alpha)$ have to be
estimated from data. For applications that require high-confidence prediction intervals where $\alpha$ is close to zero, extreme quantile 
regression methods as discussed above should be applied.
Because of model misspecification and estimation uncertainty,
the coverage in Equation~\ref{marginal_cov} does no hold in general.
The field of conformal inference \citep{Vovk2005} studies the construction of prediction intervals $\hat C_\alpha(\bm x)$ with finite-sample coverage guarantees via suitable scores, where methods are typically calibrated on a sample from $(\bm X, Y)$ of size $n_c$. Again, for high-confidence where $\alpha < 1/(n_c + 1)$, classical conformal prediction fails to guarantee correct coverage since it requires extrapolation beyond
the range of the validation sample. \cite{Pasche2025a} propose
a method based on a GPD approximation of the calibration score distribution
that, under certain assumptions, has asymptotic coverage for high-confidence prediction intervals.

\subsubsection{Supervised dimension reduction for extreme responses}\label{sec:dimred_extremeResponse}

 Another approach of regression of extreme responses with high dimensional predictors are dimension reduction methods. 
 Several tools from modern high dimensional statistics
 have recently been adapted to the extreme value context.

\cite{Gardes2018a} and \cite{Aghbalou2024} develop a dimension reduction methods for the predictor $\bm X$ with the goal to 
preserve all relevant information on the 
tail of $Y$ given $\bm X$.
To this end, they consider some orthogonal projection
$\Pi:\mathbb R^p \to \mathbb R^p$
and assume the tail conditional independence condition \citep{Aghbalou2024}
 $$ \frac{\EE |\mathbb P(Y> y \mid \bm X) - \mathbb P(Y > y \mid \Pi \bm X) |}{\mathbb P(Y > y)} \to 0, \quad y \to \infty;$$ 
see~\cite{Gardes2018a} for a different but closely related definition. The image of $\Pi$ is called a tail dimension reduction (TDR) subspace.

Under additional regularity assumptions, \cite{Gardes2018a} prove that a conditional extreme quantile estimator using the reduced predictor $\Pi \mb x$ instead of $\mb x$ is asymptotically consistent, assuming that $\Pi$ is known. %
The author proposes an algorithm to estimate the TDR space, which is suitable for moderate-dimensional settings. However, its computational complexity limits its applicability in higher dimensions, and the estimation error for the TDR  is not explicitly controlled in the theory developed in this work.

In contrast, \cite{Aghbalou2024} focuses on asymptotic theory for
an estimator of the TDR space
using an inverse regression strategy inspired by \cite{Li1991} and \cite{Cook1991}.
This classical method relies  on the fact that, under conditional independence of $Y$ and $\mb X$ given $\Pi \mb X$  and 
symmetry assumptions on the distribution of $\bm X$ given $Y$, the conditional first moment $\EE[\bm X\mid Y]$ belongs to the dimension reduction space given by the image of $\Pi$.
They adapt this framework to the extreme value setting by considering first and second moments of $\bm X$
conditional on the largest observed responses~$Y$.

While the above works consider equivalence of tail of the  conditional distributions, an arguably weaker condition is to require 
only that the tail indices of the $Y$ conditioned on $\bm X$ and $\Pi \bm X$, respectively, coincide for some projection $\Pi$. Under this assumption, \cite{Gardes2025a} estimate the conditional tail index 
by searching a matrix $B\in \rset^{p\times q}$ with $q\le p$ whose columns span the so-called tail index dimension reduction (TIDR) space, such that the conditional tail index $\gamma(\bm x)$ of $Y$ given $\bm X= \bm x$ depends only on $B^\top \bm x$. A key quantity is $\gamma_{B}(x) = \max_{\bm z \in\mathbb R^p:B^\top \bm z = B^\top \bm x} \gamma(\bm z)$. If  $\text{span}(B)$ is a TIDR space, they show that  $B$ minimizes the function $\gamma_B(\bm x)$ for all $\bm x$ over orthogonal matrices of a given rank.  Estimators of $B$ are proposed based on empirical risk minimization with consistency guarantees.

In a related line of work, \cite{Bousebata2023}
studies single-index inverse regression model $Y = g(\beta^\top \mb X) + \epsilon$, $\beta\in\rset^p$, to find linear combinations $\beta^\top \mb X$ of predictor that best explain the extreme values of the response $Y$, where $g$ is assumed to be regular varying function and 
$\epsilon$ must have a sufficiently light tail. Their estimator
maximizes the empirical covariance between $\beta^\top \bm X$ and $Y$ conditionally on $Y$ exceeding a high threshold.
The authors establish asymptotic normality of the estimated vector $\hat \beta$ and apply the method to identify key drivers of extreme cereal yields. 
\cite{Girard2025a} generalize the framework to handle weak temporal dependence and certain missing data.
In \cite{Arbel2024}, this framework is enriched by integrating a Bayesian approach, which is particularly advantageous for regularization. This extension facilitates the derivation of sparse solutions for the direction vector $\beta$, especially in high-dimensional settings and under low signal-to-noise ratios.
Moreover, \cite{Girard2024} extend the single-index model to the functional data setting, where the predictor is a possibly infinite-dimensional Hilbert space.

\subsubsection{Unbalanced classification and prediction of rare events}

In many applications, interest is in the prediction of the occurrence 
of a rare event $Y^{(u)} := \un\{Y > u\}$ given the 
predictor $\bm X = \bm x$, where $u\in\mathbb R$ is a high threshold.
Note that the event $Y^{(u)}$ is rare in an unconditional sense, i.e.,
the probability $\mathbb P( Y^{(u)} = 1)$ is very small, but the 
conditional probability $\mathbb P( Y^{(u)} = 1 \mid \bm X = \bm x)$ can be 
large. It therefore closely relates to the problem of unbalanced classification where the proportion of the minority class tends to zero.
This is in contrast to extreme quantile regression in the previous 
section, where the target is the tail of the conditional distribution
$Y\mid \bm X = \bm x$.

In this (extreme) unbalanced classification framework of rare event prediction,
\cite{Legrand2025} propose a new risk function for a family of predictors $g = (g^{(u)}, u>0)$
\begin{equation*}
R^{(u)}(g) = \frac{\PP[Y^{(u)} \neq g^{(u)}(\mb X)]}{\PP[Y^{(u)} = +1 \text{ or } g^{(u)}(\mb X) = +1]},
\end{equation*}
that penalizes overly optimistic or pessimistic classifiers $g^{(u)}:\mathbb R^p \to \{-1,+1\}$. They argue that good classifiers $g^{(u)}$ with respect to 
the limiting risk $R(g) = \lim_{u\to\infty} R^{(u)}(g)\in [0,1]$
necessarily should exhibit asymptotic dependence as in Equation~\ref{eq:chi} between the random variables $Y^{(u)}$ and the predictions $g^{(u)}(\mb X)$ as $u \to \infty$.
In cases where no classifier with such asymptotic dependence exists, 
they also consider a refined risk measure adapted to the asymptotic independence 
setting.
While they only consider linear classifiers in their theory, the proof techniques in \cite{Aghbalou2024b}, which are tailored to the minimization of a re-balanced version of the $0$--$1$ loss \citep{Menon2020}, %
could be adapted to establish non-asymptotic error bounds in their framework for more general model classes.

Similarly to \cite{Legrand2025} but in the context of time series analysis, \cite{Verma2026} define the optimal extremal classifier as the one attaining the highest extremal correlation with rare event $Y^{(u)}$.
For finite $u$, they derive a Neyman--Pearson-type characterization of optimal extreme event predictors using density ratios, and apply the methodology to 
solar flare forecasting. \cite{deCarvalho2025} study the prediction of cascading rare events
where the event $Y^{(u)}$ might trigger another extreme event $I^{(u)}$.
They develop inference methods within a flexible Kolmogorov--Arnold neural network framework.

\section{Unsupervised Settings}\label{sec:unsupervised}

\subsection{Generative AI}\label{sec_genAI}

\begin{figure}[t]
    \centering
    \includegraphics[scale=0.5]{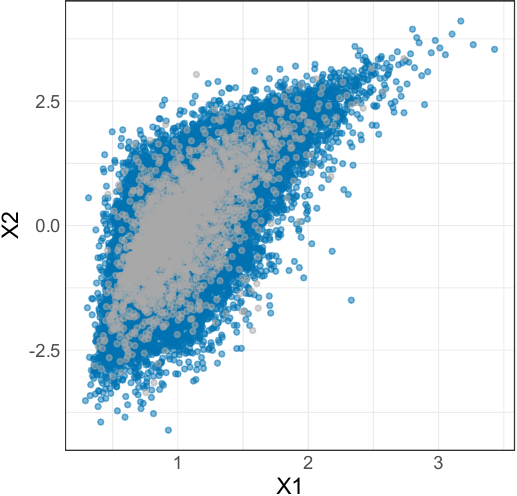}
    \includegraphics[scale=0.5]{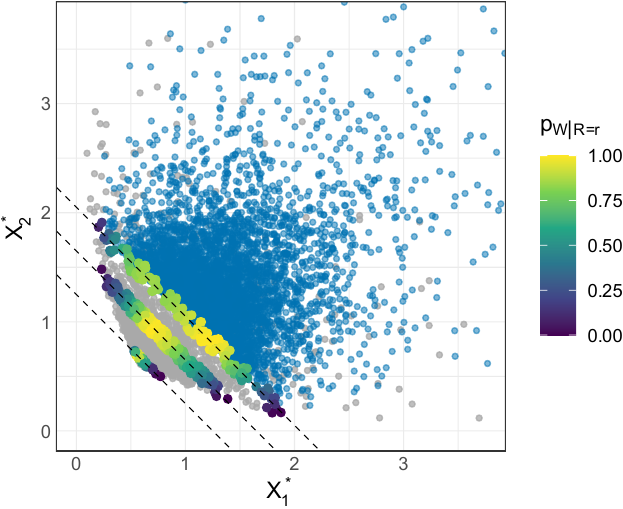}
    \includegraphics[scale=0.5]{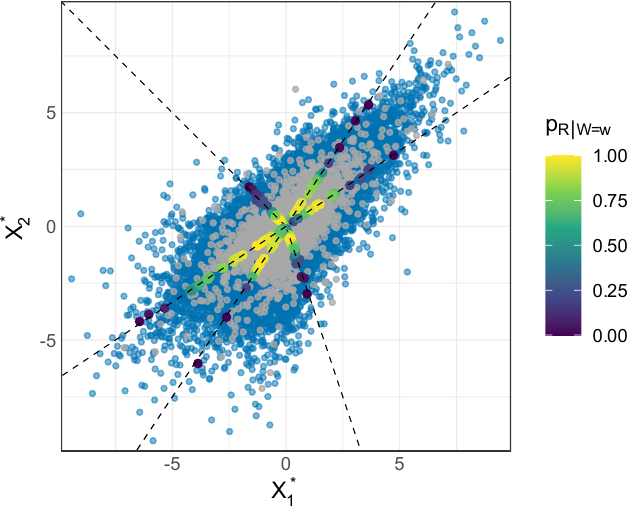}
    \caption{Right: Scatter of a random vector $\bm X$ on the original scale.
    Center: Scatter of the transformed $\bm X^*$ on Pareto margins, up to radial rescaling for visual clarity. The generative model learns the distribution of the radius $p_R$, and conditional on $R = r$, the angular distribution $p_{\bm W \mid R = r}$.
    Left: Scatter of the transformed $\bm X^*$ on Laplace margins. The generative model learns the distribution of the angle $p_{\bm W}$, and conditional on $\bm W = \bm w$, the radial distribution $p_{R \mid \bm W = \bm w}$.
    Grey points correspond to training observations, while blue points correspond to observations sampled by the generative models.
    }
    \label{fig:gen-ai}
\end{figure}

Generative artificial intelligence (GenAI) refers to a class of machine learning models that aim to learn the probability distribution of a complex, high-dimensional random vector $\bm X\in\mathbb R^p$ and can generate new, realistic samples from it;
see the left-hand side of Figure~\ref{fig:gen-ai} for a bivariate example. The state-of-the-art methods have proved highly effective for this task and are now standard tools in applications
ranging from language modeling \citep{Vaswani2017} to image generation \citep{Goodfellow2014}, as well as AI weather forecasting \citep{Price2024}.

Most of the approaches rely on a generator $G: \mathbb R^q \to \mathbb R^p$ that transforms 
a set of latent variables $\bm Z\in\mathbb R^q$ to have the desired distribution of $\bm X$, i.e.,
\begin{align}
    \label{generator}
    \bm X \stackrel{d}{=} G(\bm Z), \qquad \bm Z \sim p_{\bm Z}
\end{align}
where the distribution $p_{\bm Z}$ is often taken to be an independent normal or uniform distribution. Generative models learn an approximation $G_\vartheta \approx G$ of the generator from the training observations $\bm X_1,\dots,\bm X_n \sim \mathcal P$, in such a way that the distribution
$p_\vartheta$ of $G_\vartheta(\bm Z)$ approximates best $p_{\bm X}$.
The discrepancy between the training sample and the learned distribution
is typically measured by some distance divergence measure, and then 
minimized using maximum likelihood or variational methods.
Once trained, sampling from $G_\vartheta(\bm Z)\sim p_\vartheta$ is cheap and  produces data that approximately follow the distribution
$p_{\bm X}$.
There are several popular models for $G_\vartheta$, including generative adversarial networks (GANs), variational autoencoders (VAEs), diffusion models, and normalizing flows.

A fundamental limitation of these methods concerns their extrapolation capabilities in terms of learning the correct distributional tail of $p_{\bm X}$. 
Extreme observations generated from 
$G_\vartheta(\bm Z)$ may not match the actual marginal distributions
and/or dependence structure of extremes of $\bm X$. 
In many applications in finance or meteorology, such events are the most 
relevant for the system.
The reason for this limitation is the lack of sufficient extreme training
data and the fact that the training loss typically focuses on the 
distributional bulk rather than the extremes.
Extreme value theory described in Section~\ref{sec:background} provides theoretically justified tools
to extrapolate in a principled way beyond the data range
and can therefore improve GenAI methods to better represent distributional tails. 

First, theoretical results show that the marginal distributions of
the generated samples from a GAN are either bounded or light-tailed if the latent input $\bm Z$ is uniform or Gaussian, respectively \citep{Wiese2019, Huster2021}. \cite{Huster2021} and \cite{Girard2024a} use heavy-tailed noise as input $\bm Z$ and show that this improves the representation of the marginal, and to some extent, the multivariate tails.
Instead of changing the latent distribution, 
\cite{McDonald2022} and \cite{Boulaguiem2022} fit GPD and GEV distributions to 
marginal distributions of the data, respectively, and then transform to
the copula scale to learn the dependence structure through a flexible generator. \cite{Allouche2022} reparametrize the 
generator to make the univariate tails learnable by neural networks. While these approaches allow extrapolation of marginal distributions beyond the data range, there is no guarantee that the extremal dependence between the components of $\bm X$ is learned correctly.
A different approach is \cite{Bhatia2021}, which uses a conditional GAN with conditioning on the extremeness of a risk functional.

Apart from the marginal distributions, the dependence
structure can differ between the bulk and the tail,
and classically trained GenAI methods may not reliably reproduce such tail dependence.
For example, \cite{Lafon2023} show that classical generative models based on feed-forward networks with ReLU activations produce angular measures that concentrate on a finite set of points.
Multivariate extreme value theory has developed a range of theoretically justified approaches to describe and learn 
multivariate tail dependence structures; see Section~\ref{sec:mevt}.
Most of these approaches rely on the fact that extremal dependence
simplifies when the data is transformed to certain standardized margins.

We first describe a general recipe for these methods and then give concrete examples.
\begin{itemize}
    \item[(i)] Transform the random vector $\bm X$ to some standardized margins $F_*$ via componentwise probability integral transforms, i.e., $\bm X^* = F_*^{-1}\{ F_{\bm X} (\bm X)\}$.
    \item[(ii)] Assume a simpler dependence structure in the tail of $\bm X^*$ that is asymptotically motivated by extreme value theory, and learn an extrapolation-aware generator $G^*_\vartheta(\bm Z)$ for this representation. Most methods exploit a radial-angular decomposition
    \begin{equation}\label{radial_angular} 
    R \coloneqq \norm{\bm X^*}, \quad \bm W \coloneqq \bm X^* / R, 
    \end{equation}
     where the choice of the norm depends on the specific method.
    Extrapolation is typically achieved by extrapolating the radius $R$.
    \item[(iii)] Generate samples $\hat{\bm X}^*$ from  $G^*_\vartheta(\bm Z)$ and transform back to original scale by $\hat{\bm X} = F_{\bm X}^{-1}\{F_*(\hat{\bm X}^*)\}$, where the model for $F_{\bm X}$ should also support extrapolation beyond the observed range.
\end{itemize}
The joint distribution of $(R, \bm W)$ in Equation~\ref{radial_angular} admits the factorization
\begin{equation}\label{eq:factorization}
    p_{R, \bm W}(r, \bm w) \stackrel{(i)}{=} p_{R}(r) p_{\bm W \mid R}(\bm w \mid r)
    \stackrel{(ii)}{=} p_{\bm W}(\bm w) p_{R \mid \bm W}(r \mid \bm w).
\end{equation}
Methods based on multivariate regular variation (MRV) use factorization $(i)$, while methods based on the geometric approach use factorization $(ii)$.

Under MRV, $\bm X^*$ is typically standardized to Pareto margins, the radial measure $p_R(r)$ is Pareto, and the angular measure $p_{\bm W \mid R}(\cdot \mid r)$ becomes independent of the radius for large values of $r$; see the center panel of Figure~\ref{fig:gen-ai}.
\cite{Lafon2023} already assume that the data are on a Pareto scale, 
and then learn two VAEs: an unconditional VAE to sample the heavy-tailed radius $R \sim p_R$, and a conditional VAE to sample the angular component $W\mid R \sim p_{\bm W \mid R}(\cdot \mid r)$, enforcing independence between the radius and angle for large $r$.
\cite{Lhaut2025} model the margins of $\bm X$ with GPD, transform the margins to Pareto scale $\bm X^*$, and train a Wasserstein GAN on $\bm X^*$ to learn the angular measure $\bm W \mid R > r$ above a high threshold $r$.
In a similar vein, \cite{Hu2025} use a normalizing flow to jointly learn the margins and the tail dependence of $\bm X$ in an MRV framework. 
Following the maxima approach, \cite{Hasan2022} fit a generative model to the Pickands dependence function of a multivariate extreme value distribution.

The MRV framework becomes uninformative if the data exhibit asymptotic independence, because the angular measure concentrates its mass on the axes. In this case, several complementary approaches have been proposed. 
Under the geometric extremes framework, the data is standardized to $\bm X^*$ with  exponential or or Laplace margins. The angular distribution $p_{\bm W}$ is first modeled, and the conditional radial density then takes the form
$$p_{R \mid \bm W} (r \mid \bm w) \propto r^{d-1} \exp[-r g(w)],$$
where large values $g(w)$ correspond to a direction $w$ with lighter tail, in the sense that it controls the exponential rate of decay of tail mass in that direction; see the right-hand side of Figure~\ref{fig:gen-ai}.
After standardization to Laplace margins, 
\cite{Murphy-Barltrop2024} use a neural network as a flexible parameterization of the gauge function.
\cite{Mackay2025} decompose the density of $\bm X$ via an angular-radial decomposition as in $(ii)$ of Equation~\ref{eq:factorization}. Their model is semi-parametric since they assume that $p_{R \mid \bm W}$ is in the domain of attraction of a $\mathrm{GPD}(\sigma(\bm w), \gamma(\bm w))$ where $(\sigma(\bm w), \gamma(\bm w))$ is fitted with a dense neural network, and $p_{\bm W}$ is learned with kernel density estimation.
\cite{Wessel2025} introduce and compare different GenAI approaches to model angular distributions as they appear in many extreme models, including multivariate regular variation, the geometric approach, and the SPAR model.
See also \cite{Allouche2026} for a more detailed review on simulation on extreme events with neural networks.

\subsection{Anomaly Detection}\label{sec:anomaly}

Anomaly detection has emerged as a critical area in data science, driven by applications in fraud detection, aviation safety, health monitoring of complex systems and food risk assessment. Anomalies are typically defined as observations that deviate significantly from the training data $\bm X_1, \dots, \bm X_n$ from distribution $\mathcal P$. For a comprehensive overview, see the classical review by \cite{Chandola2009} and more recent works focusing on deep learning approaches \cite{Pang2021}.

Quite early, it has been realized that the detection of anomalies
can be phrased as an extreme value theory problem. The general idea is that a point $\bm x_0$ of interest 
should be labeled as an anomaly (or novelty), if it lies 
in a very low-probability region of the training distribution $\mathcal P$.
In order to accurately quantify this probability, the extreme 
value approximations discussed in Section~\ref{sec:background} 
are used. We discuss the literature that uses univariate 
approximations to improve anomaly detection in Section~\ref{sec:uni_anom}. Methods that aim to 
classify extreme points as anomalies based on multivariate
extreme value theory are considered in Section~\ref{sec:mult_anom}.

\subsubsection{Extreme value theory for anomaly detection}
\label{sec:uni_anom}

\cite{Roberts1999} fit a Gaussian mixture model
to the training data and use the GEV distribution to 
classify a new point as normal or abnormal; see
also \cite{Clifton2011} for an extension to
multivariate, multimodal distributions.
In the context of time series analysis,  
\cite{Siffer2017} fit a GPD to the tail of univariate data
to detect anomalies and describe an algorithm for online updates 
of the decision threshold.

For novelty detection in a multi-class classification setting, the extreme value machine in \cite{Rudd2018} fits a GEV distribution to the smallest distance of a training point $\bm X_i$ to any point belonging
to a different class. If a new point $\bm x_0$ is not close enough
to any of the training points, measured by the probabilities of the respective GEV, it is classified as belonging to a novel, unseen class.
\cite{Vignotto2020} discuss theoretical shortcomings of this approach
in terms of how extreme value theory is applied, and 
propose the GPD classifier as alternative method for anomaly detection.
Under the null hypothesis of a normal point $\bm x_0\in\mathbb R^p$, they show that the lower tail of the distribution of distances $D_i = \| \bm X_i - \bm x_0 \|_2$ belongs to the min-domain of attraction of a GEV distribution with shape parameter $\gamma = -1/p$, since for small $\delta >0$
\begin{align}\label{dist_approx}
\mathbb P( D_i < \delta ) \approx f(\bm x_0) V_p(\delta) = f(\bm x_0) C_p \delta^p, 
\end{align}
where $f(\bm x_0)$ is the training density at $\bm x_0$ under $\mathcal P$, $V_p(\delta)$ is the volume of ball of radius $\delta$ in $\mathbb R^p$, and $C_p>0$ is a constant depending only on the dimension $p$.
Under the alternative hypothesis of an anomaly from a distribution with non-overlapping support, the Hill estimator converges almost surely to $0$, enabling the construction of a statistical test for anomaly detection.

\begin{textbox}{Intrinsic dimensionality in machine learning}
    Similar methods as for anomaly detection based on extreme value theory are used in the estimation of the intrinsic dimensionality, which describes the effective dimension of a high-dimensional data set \citep{Amsaleg2015}. In this case,
    the lower tail of the distances in Equation~\ref{dist_approx}
    is driven by the intrinsic dimension $p'<p$.    
     Intrinsic dimensionality of the predictor distribution can be connected to key properties of machine learning algorithms, such as
    the susceptibility to adversarial attacks \citep{Bailey2022}.
\end{textbox}

\subsubsection{Anomaly detection for multivariate extremes}
\label{sec:mult_anom}

A second line of work focuses on anomaly detection for extreme multivariate observations, aiming to distinguish ``normal'' from ``abnormal'' extremes. This line of work assumes that anomalies in the bulk are handled separately.
These methods consider the setting where the vector $\bm X$ standardized to Pareto margins is multivariate regularly varying and define the anomaly score for observation $\bm x$ as
$$
s(\bm x) = \text{radial score(}\bm x) \times \text{ directional score(} \bm x),
$$
where radial score($\bm x)$ is low if $\bm x$ has large radius, while directional score($\bm x)$ is low if $\bm x$ lies along ``unusual directions''. 

In high dimensions, 
\cite{Goix2016,Goix2017} partition the upper tail of $\bm X$ into cone-type regions which represent different configurations of joint extremes.
Their directional anomaly score is defined as the empirical frequency of training points in each cone-type region, yielding a sparse description of normal extremal behavior.
This framework has also proven %
useful for clustering anomalous extreme events \citep{Chiapino2020}. %

In low-dimensional settings, instead of partitioning the upper tail of $\bm X$ into cone-type regions, \cite{Thomas2017}  directly model the angular structure through angular mass-volume sets \citep{Scott2005}.
A mass-volume set of mass $\alpha$ for $\bm X$ is the smallest (in terms of Lebesgue measure) set $A \subseteq \mathbb{R}^p$ with probability at least $\alpha$. 
\cite{Thomas2017} estimate mass-volume sets of high mass $\alpha$ for the angular component $\bm W$ of $\bm X$. These sets correspond to the most likely directions where extreme points occur.

In practice,  a finite grid of values $[\alpha_1, \ldots, \alpha_M]$ is chosen, yielding a family of nested mass-volume sets $\Omega_1\subset\ldots \subset\Omega_M$ constructed by constrained optimization of an empirical criterion, and the directional score of a new angle $\mb w$ is a stepwise function which is constant on the set differences $\Omega_{m+1}\setminus \Omega_{m}$. 

Theoretical refinements for empirical standardization into Pareto margins are provided in \cite{Clemencon2023}. The concept of mass-volume sets also plays a central role in \cite{Cai2011}, where it is employed to delineate multivariate quantile contours. While their primary objective differs from that of anomaly detection, the two tasks of quantile contour estimation and anomaly detection are closely related, as both rely on characterizing the extreme regions of the data distribution.

\subsection{Sparsity and Dimension Reduction}\label{sec:dimRed}

The period from roughly 2015 to 2020 has marked the beginning of a movement bridging extreme value theory with methods for sparsity and high dimensions from statistical learning.
\cite{Engelke2021} reviewed the literature at that time with a 
focus on techniques for dimension reduction such as PCA, %
clustering %
concomitant extremes %
and graphical modeling. %
Research in these fields has made significant progress in recent years, and we briefly review recent works.

Following earlier clustering approaches for extreme observation \citep{Chautru2015, Chiapino2019, Janssen2020},  new methods have been developed
using spherical $k$-principal-components clustering \citep{Fomichov2023}, spectral clustering \cite{Medina2024} or kernel PCA on the angular measure \cite{Medina2025}. Clustering based on maxima in time series is studied in \cite{Boulin2025a}, with applications to clustering precipitation records over Europe \citep{Boulin2025b}. Another approach involves the use of latent linear factor models for dimension reduction of the tail distribution \cite{Boulin2026}.
The previously open questions of dimension selection for PCA and subspace identification \citep{Cooley2019,Drees2021} has been addressed in \cite{Butsch2025b,Butsch2025a} and further applications have been proposed \citep{Rohrbeck2023}. Regular variation and functional extensions of PCA have been developed for observations in separable Hilbert spaces with applications to real data \citep{Clemencon2024}.
PCA for max-linear models establishes connections with tropical algebra \citep{Reinbott2026}. On exponential margins, representations of extremes on a hyperplane have been proposed to facilitate modeling and linear dimension reduction, such as PCA \citep{Wan2026}.

The theory and methodology for graphical modeling of extreme values \citep{Gissibl2018, Engelke2020, Segers2020} has been further extended in several directions.
There are now structure learning methods with high-dimensional
recovery guarantees for tree models \citep{Engelke2022} by
minimum spanning trees, and for general graphs in the class of H\"usler--Reiss distributions through
lasso-type $L_1$-penalization \citep{Engelke2025e, Wan2025}
and with latent variables \cite{Engelke2025c}.
In the same model class, efficient statistical inference techniques exploit matrix completions 
\citep{Hentschel2025}, score matching \cite{Lederer2024}, and convex optimization problems with positive dependence \citep{Rottger2023} and more general constraints \citep{Echave-SustaetaRodriguez2025}.
In the context of max-linear graphical models, \cite{Tran2024} estimate directed tree structures.

Directed extremal graphical models have
been studied in connection with structural causal models and 
extremal treatment effects \citep{Engelke2025d, Bai2026}.
Asymptotically independent data can be modeled in a sparse way by 
extremal graphical models based on the generalized notion of conditional independence in \cite{Engelke2025} or the geometric approach in \cite{Papastathopoulos2026}.
Finally, a stochastic partial differential equation framework 
for sparse spatial extremes has been developed  for the new
class of intrinsic Whittle--Mat\'ern Brown--Resnick fields
\citep{Bolin2025}. For a more complete review on topics related to extremal
graphical models we refer to~\citep{Engelke2024}.

\section{Concluding Remarks}\label{sec:conclusion}

\begin{issues}{Future Issues}
\begin{enumerate}

\item The current extrapolation methods for supervised learning make simplifying assumptions, such as additive models or multivariate regular variation. Overcoming such regularity assumptions and integrating the underlying principles 
into large-scale machine learning methods, such as neural networks for AI weather modeling, is an important future research direction.
\item Evaluation of extrapolation performance is inherently difficult, which also %
deters fair model comparison and data-driven model selection. Especially for flexible models without rigorous statistical guarantees, a robust evaluation framework will be crucial in the future.    
\item Asymptotic and non-asymptotic statistical theory for extreme value estimators is currently mostly limited to approaches based 
on multivariate regular variation. Extending this theory to approaches for asymptotic independence would further increase confidence in predictions from these methods.
\item Many approaches in this review adapt machine learning 
tools to the framework of extreme value theory. While their
empirical effectiveness is checked on simulations, theoretical guarantees are still missing in some cases and require non-trivial adaptations
of proofs in classical learning theory.
\item Other fields such as causality or distributional regression have recently produced new approaches to extrapolation. A combination of these methods with tools based on extreme value theory discussed in this review could turn out to be fruitful in the future.
\end{enumerate}
\end{issues}

Extrapolation is a hard problem and any method from statistical
learning has to rely on certain assumptions. 
Extreme value theory provides one avenue for creating 
such assumptions that naturally arise through limiting 
theorems for the tails of univariate and multivariate 
distributions. 
Other types of assumptions for extrapolation exist in
fields such as causality \citep{Rothenhausler2021} or distributional regression 
\citep{Shen2024}.
In general, there will not be one universally best method
for all problems, but the suitability of an approach will rather  depend on the 
domain of application and the structure of the underlying
data generating process.
In future research, it is therefore crucial to have different frameworks and clearly stated sets of assumptions that cover
different scenarios (e.g., asymptotically dependent versus independent multivariate data).

Evaluation of the extrapolation performance of predictions or generated distributions is
extremely difficult in practice. The reason is
that the region that extrapolation targets has by definition
few data points, and therefore classical cross-validation 
is not applicable. Future research 
should develop robust evaluation frameworks for model comparison and selection that are particularly tailored to tail regions
of predictor or response distributions.
Given this inherent difficulty of model evaluation, 
the theoretical analysis of new models should play 
a primary role in future work. Indeed, such results
would highlight in which situation a model can be used
with confidence and point at possible limitations.

\bibliographystyle{abbrvnat-arxiv}
\bibliography{Review_Article.bib}
\end{document}